\pdfoutput=1
\documentclass[11pt]{article}

\usepackage{acl}

\usepackage{times}
\usepackage{latexsym}

\usepackage[T1]{fontenc}
\usepackage[utf8]{inputenc}
 
\usepackage{microtype}

\usepackage{amsmath}
\usepackage{amsfonts}       % blackboard math symbols
\usepackage{graphicx}
\usepackage{tabularx}
\usepackage{multirow}
\usepackage{booktabs}
\usepackage{subcaption}
\usepackage{tikz}
\usepackage{mathtools}
\usepackage{soul}
\usetikzlibrary{arrows,automata}
\usepackage{amsthm}

\usepackage{arydshln}
\usepackage{xcolor}

\def\arraystretch{1}%  1 is the default, change whatever you need
\newcommand{\ra}[1]{\renewcommand{\arraystretch}{#1}}

\usepackage[disable]{todonotes}
\usepackage{lipsum}
\usepackage{tabularx}

\title{Why Exposure Bias Matters: An Imitation Learning Perspective of Error Accumulation in Language Generation}

\author{Kushal Arora$^1$\thanks{\ \ A part of this work was done when the author was an intern at Borealis AI.}
 \quad Layla El Asri$^2$ 
 \quad Hareesh Bahuleyan$^3$\thanks{\ \ During a part of this work, the author was an employee at Borealis AI.}
 \quad Jackie Chi Kit Cheung$^1$\thanks{\ \ During a part of this work, the author was an Academic Advisor at Borealis AI.} \\ \\
     $^1$Mila / McGill University \quad $^2$Borealis AI \quad
     $^3$ Zalando SE \\
     {\small \{\tt kushal.arora@mail, jcheung@cs\}.mcgill.ca} \\
     \small \tt layla.elasri@borealisai.com, hareeshbahuleyan@gmail.com   
}

\begin{document}
\maketitle
\begin{abstract}
  Current language generation models suffer from issues such as repetition, incoherence, and hallucinations. An often-repeated hypothesis is that this brittleness of generation models is caused by the training and the generation procedure mismatch, also referred to as exposure bias. In this paper, we verify this hypothesis by analyzing exposure bias from an imitation learning perspective. We show that exposure bias leads to an accumulation of errors, analyze why perplexity fails to capture this accumulation, and empirically show that this accumulation results in poor generation quality. \footnote{Source code to reproduce these experiments is available at \url{https://github.com/kushalarora/quantifying_exposure_bias}.}
\end{abstract}

\section{Introduction}
Large-scale neural language models have made great strides in language generation tasks such as machine translation~\cite{bahdanau_neural_2014,vaswani_attention_2017, raffelExploringLimitsTransfer}, text summarization~\cite{see_get_2017,lewis_bart_2019, zhang2019pegasus}, conversational dialog generation~\cite{serban_building_2015,lowe_towards_2017, rollerRecipesBuildingOpendomain2020, zhangDIALOGPTLargeScaleGenerative2020}, etc. 

However, despite the successes achieved by these models on several conditional generation tasks, they continue to suffer from degenerate behaviors such as repetition, a lack of diversity, dullness, and incoherence, especially in open-ended generation settings such as text completion and dialog modeling~\cite{holtzman_curious_2019}. This degenerate behavior is often attributed to a mismatch between the \textit{maximum likelihood training} and generation procedure mismatch~\cite{welleck_neural_2019,choi2SoftmaxDiversifyingNeural2020,liDiversityPromotingObjectiveFunction2016}.

% However, despite the successes achieved by these models on several conditional generation tasks, they continue to suffer from degenerate behaviors such as repetition, a lack of diversity, dullness, and, incoherence. This behavior is especially acute while using deterministic decoding schemes such as greedy search in open-ended generation tasks such as text completion and dialog modeling~\cite{holtzman_curious_2019}. 
% In the literature, this degenerate behavior is often attributed to a mismatch between the maximum likelihood training and generation procedure mismatch~\cite{welleck_neural_2019,choi2SoftmaxDiversifyingNeural2020,liDiversityPromotingObjectiveFunction2016}.

% Numerous solutions have been proposed to address natural language degeneration. These methods either tweak the training objective to disincentivize degenerate behavior, such as unlikelihood training~\cite{welleck_neural_2019}, or employ smarter decoding strategies such as repetition penalties~\cite{keskarCTRLConditionalTransformer2019}, sampling with temperature~\cite{caccia_language_2018}, top-k, and nucleus sampling~\cite{holtzman_curious_2019}. These methods have addressed some of the concerns regarding dull language generation resulting in better generation quality according to both the automatic metrics and the human judgment scores.

%Language generation models are generally trained using maximum likelihood estimation (MLE), i.e., by maximizing the likelihood of the training corpus. This is also referred to as teacher forcing~\cite{williams_learning_1989}.

Maximum likelihood training, also referred to as \textit{teacher forcing}~\cite{williams_learning_1989}, factorizes the language model as a linear chain, and maximizes the log-likelihood of this factorized language model on a training corpus. During this maximum likelihood training, the model learns a distribution of the next tokens conditioned on the contexts from the ground-truth training data. 

A concern with MLE-based training is that ground-truth contexts from the training corpus are not available during generation. Rather, the conditioning contexts during this phase comprise tokens previously generated by the model itself. The distribution of these contexts seen during the generation phase might be very different from the ones encountered during the training phase. This mismatch is referred to as \textit{exposure bias}~\cite{ranzato_sequence_2015,bengio_scheduled_2015}.

A side effect of exposure bias is that an error at any step during generation might have a cascading effect as the next context will incorporate this erroneous prediction, deviating away from the ground truth context distribution, leading to more errors. Several authors~\cite{welleck_neural_2019,choi2SoftmaxDiversifyingNeural2020,liDiversityPromotingObjectiveFunction2016} have speculated that these errors might result in sequences that degenerate over the sequence length resulting in incoherent text, lack of vocabulary diversity, and hallucinations, and word- and phrase-level repetition.

There is an active debate in the language generation community on the impact of exposure bias in language generation. Authors have both validated~\cite{xu_rethinking_2020,zhang_bridging_2019} and questioned~\cite{he_quantifying_2020} the impact of exposure bias on language generation. Previous works have also linked exposure bias to out-of-distribution generalization~\cite{schmidt_generalization_2019}, and out-of-domain generalization and hallucinations~\cite{wang_exposure_2020} but these claims remain weak in absence of a clear and principled formalization of the exposure bias issue. Finally, several approaches have been proposed to mitigate exposure bias~\citep{ranzato_sequence_2015,shen_minimum_2016,bahdanauActorCritic2017,chen_reinforcement_2020,leblond_searnn:_2017, welleck_neural_2019}, though these approaches improve the performance on the downstream tasks, the authors neither formalized exposure bias nor provided any empirical evidence that the downstream improvements are directly linked to the mitigation of exposure bias issue. 

In this paper, we attempt to clarify this confusion by formalizing exposure bias in terms of accumulation of errors and by analyzing its impact on generation quality. We do this by providing a theoretically grounded understanding of the exposure bias issue by analyzing it from an imitation learning perspective. We use this perspective to show that behavior cloning---an imitation learning algorithm---is equivalent to teacher forcing under the choice of a particular loss function. We then exploit this equivalence by borrowing the bound on error accumulation caused by behavior cloning and use it to formalize exposure bias and analyze error accumulation in language generation.

 Finally, we use this quantifiable definition of exposure bias to demonstrate that models trained using teacher forcing do suffer from an accumulation of errors. We also show, both analytically and empirically, why perplexity fails to capture this error accumulation, and how a lower exposure bias correlates with better generation quality.

%  Finally, we measure the impact of various decoding schemes on mitigating the effect of exposure bias, and show, both qualitatively and quantitatively, that a lower exposure bias results in better generation quality.

\section{Language Generation Formulation}
Given a finite-sized vocabulary set $\mathcal{V}$, language generation is posed as a problem of generating a variable-length sequence $w_{0}^{n}\in \mathcal{V}^n$ from a language model $p_{\theta}$, either unconditionally or conditioned on a source $\bf{x}$, using a decoding algorithm $\mathcal{F}$: 
\begin{equation}
  w_{0}^{n} = \mathcal{F}(p_{\theta}; \bf{x})
\end{equation}
Language modeling is the problem of learning this parameterized model $p_{\theta}$ that approximates an oracle model $o$. 
% such that decoding from the model $p_{\theta}$ mimics greedily sampling from the oracle $o$. 

Maximum likelihood-based training factorizes the probability distribution model, $p_{\theta}(w_{0}^{n})$, into a linear chain, i.e.,
\begin{equation} \label{eq:lang_modeling_def} 
    p_{\theta}(w_{0}^{n}; \textbf{x})=\prod_{i=1}^{n}p_{\theta}(w_i|w_{0}^{i-1}; \textbf{x})p(w_0),
\end{equation}
where $w_i$ is the token to be generated at step $i$ and $w_{0}^{i-1}$ is the context at time $i$; i.e., all the tokens seen from step 0 to step $i-1$.\footnote{As $w_0$ is usually a fixed SOS token, $p(w_0) = 1$. We will drop $p(w_0)$ from the subsequent equations for brevity.}

During maximum likelihood training, the language model is trained by minimizing the negative log-likelihood on the corpus $\mathcal{D}$, i.e., 
\begin{equation} \label{eq:tf_lm_loss_def} 
  \theta^{*} = \underset{\theta}{\text{argmin}} \frac{-1}{|\mathcal{D}|} \sum_{w_{0}^{n} \in \mathcal{D}} \sum_{i=0}^n \log p_{\theta}(w_{i}|w_{0}^{i-1}),
 \end{equation}
 where $|\mathcal{D}|$ is the number of tokens in the corpus. 

Given a trained language model $p_{\theta}$, the simplest strategy for generating a target sequence is to greedily sample the model;i.e., at each step $i$, pick the most probable token $w_i = \arg\max p_{\theta}(\cdot|w_{0}^{i-1}; x)$---as its prediction. For the next step $i+1$, we use $w_i$ to generate the context $w_{0}^{i} = w_{0}^{i-1}w_i$, and use it to predict the next token. This continues either until the maximum sequence length ($T$) is reached, or a special end-of-sequence token ($\text{EOS}$) is generated.  
%In our analysis, unless stated otherwise, we will assume we decode using a greedy search strategy.~\footnote{This is a reasonable choice. Ideally, if $p_{\theta} = o$, greedy sampling will be the optimal strategy to generate the target sequence conditioned on the input $\bf{x}$.}

\section{An Imitation Learning Perspective of Language Generation} \label{sec:lg_as_il}
In this section, we will present an imitation learning perspective of language generation. This framing will allow us to borrow theoretical machinery from the imitation learning literature to formalize the exposure bias issue and analyze it in terms of the accumulation of errors due to a procedural mismatch between MLE-based training and generation.

We start by posing language generation as a sequential decision-making problem and language modeling as an instance of imitation learning. We exploit these parallels to show behavior cloning, an imitation learning algorithm, is equivalent to teacher forcing under a choice of a particular loss function. We then exploit this equivalence to quantify the error accumulation due to exposure bias.

\paragraph{Language Generation is a Sequential Decision-Making Problem:}
A sequential decision-making problem can be formalized as learning a policy $\pi(a_{t} | s_{t})$ over a space of actions $a_t \in \mathcal{A}$ and states $s_t \in S$ where the next state $s_{t+1}$ is conditioned on the current state-action pair and is determined by the transition distribution $P(s_{t+1}| s_{t}, a_{t})$. 
We can use this framework to pose language generation as an instance of a sequential decision-making problem with language model $p_{\theta}$ as the policy, contexts $w_{0}^{t-1} \in \mathcal{V}*$ as states, the next token prediction $w_t \in \mathcal{V}$ as actions, and concatenation as the transition function.

This perspective allows us to appreciate the fact that, during generation, predictions at previous steps affect the next predictions, and errors over time can cascade resulting in incoherent sequences.

% The language generation with greedy decoding leads to the prediction of the next word $w_{t}$ being conditioned on the previous decisions made while predicting $w_{i|0 < i \le t-1}$.

\paragraph{Language Modeling is Imitation Learning:} \label{sec:lm_as_il}
Imitation learning is a class of methods to solve a sequential decision-making problem while having access to the oracle policy $o$ or data generated by the oracle; i.e., $\mathcal{D} = \{(s_t, a_{t}) | s_{t} \sim d^{t}_{o}, a_{t} \sim o(\cdot | s_t)\}$. Here, $d_{o}^t$ is the oracle-induced state-visitation distribution at time $t$. 

In imitation learning, an agent learns a model policy $\pi$ that reproduces the expert policy $o$ but on the state-visitation distribution $d_{\pi}^t$ that has been induced by the model policy $\pi$, i.e.:
\begin{equation*}
  \pi^{*} =  \underset{\pi}{\arg \min}\ \sum_{t=1}^{T} \mathbb{E}_{s \sim d_{\pi}^t} \Big[ l(\pi, s; o) \Big].
\end{equation*}
where $l(\pi, s; o)$ is the expected per-step cost of  model $\pi$ mimicing oracle $o$ in state $s$, $d_{\pi}^t$ is the state-visitation distribution at step $t$ induced by following policy $\pi$ from step $1$ to $t-1$.
 
The sequential decision-making perspective of language generation allows us to pose language modeling as an instance of imitation learning---learning a model for a sequential decision-making problem with the help of an expert oracle (in RL-based methods) or using the data generated by the oracle (for MLE-based methods).

\paragraph{Teacher Forcing is Behavior Cloning:}~\label{sec:tf_as_bc}
The assumption of access to an oracle is unrealistic in many scenarios. Behavior cloning is an approach to solving an imitation learning problem using only the training data generated by an oracle. In this setup, the state-action pairs in the training data are assumed to be identically and independently distributed. This is equivalent to reducing a sequential decision-making problem to a supervised multi-class classification learning problem. 

Concretely, this learning problem can be seen as minimizing the expected per-step loss under the state distribution induced by the oracle:
\vspace{-1em}
\begin{align} \label{eq:behv_cloning_loss_def}
  \hspace*{-1em} L^{BC}(\pi) =& \sum_{t=1}^{T} \mathbb{E}_{s_t \sim d_{o}^t} \left[l(s_t,\pi;o)\right] \\
                       \approx& \frac{-1}{|\mathcal{D}|} \sum_{(s_t, a_t) \in \mathcal{D}} l(s_t, \pi;o),
\end{align}
Here, $ L^{BC}(\pi)$ is the behavior cloning loss, and $l(s,\pi; o)$ is the per-step loss. 

Similarly, in practical scenarios, language models are also trained on a finite training corpus, $\mathcal{D}$, that is assumed to be generated by the oracle; i.e., $\mathcal{D} = \{(w_0^{t-1}, w_{t})| w_0^{t-1} \sim d_{o}^{t-1}, w_{t} \sim o(\cdot|w_0^{t-1})\}$.

The maximum likelihood training loss from Equation \ref{eq:tf_lm_loss_def}, can be reformulated as learning the distribution over the next tokens, conditioned on the training contexts generated by the oracle, $w_{0}^{t-1} \sim d_{o}^{t-1}$:
\begin{align} 
  \hspace{-0.5em} L^{\text{TF}}(p_{\theta}) = \frac{-1}{|\mathcal{D}|} \sum_{(w_{0}^{i-1}, w_i) \in \mathcal{D}} \log p_{\theta}(w_{i}|w_{0}^{i-1})&,\label{eq:tf_loss_def_1} \\
                  \approx \sum_{t=1}^{T} \underset{\substack{w_{0}^{t-1} \sim d_{o}^t \\ w_{t} \sim o(\cdot|w_0^{t-1})}}{\mathbb{E}} \left[ -\log p_{\theta}(w_{t}|w_{0}^{t-1}) \right]&, \label{eq:tf_loss_def_2}
\end{align}
The behavior cloning loss (Equation \ref{eq:behv_cloning_loss_def}) is equivalent to the language modeling loss (Equation \ref{eq:tf_loss_def_2}) with $l(p_{\theta}, w_{0}^{t-1}; o)$ defined as,
\begin{equation*}
  l(p_{\theta}, w_{0}^{t-1}; o) = \underset{w_{t} \sim o(\cdot|w_0^{t-1})}{\mathbb{E}}  \left[-\log p_{\theta}(w_{t}|w_{0}^{t-1})\right]
\end{equation*}

% The standard approach for training a language model is by minimizing the negative log-likelihood of the training corpus. The loss function for training a language model is:
% \begin{align} 
%  L(p_{\theta})  & = \frac{-1}{|\mathcal{D}|} \sum_{(w_{0}^{i-1}, w_i) \in \mathcal{D}} \log p_{\theta}(w_{i}|w_{0}^{i-1}),\label{eq:tf_loss_def_1}, \\ 
%        & = \frac{-1}{|\mathcal{D}|} \sum_{w_{0}^{n} \in \mathcal{D}} \sum_{i=0}^n \log p_{\theta}(w_{i}|w_{0}^{i-1}),
% \end{align}
% where $|\mathcal{D}|$ is the number of tokens in the corpus. 

For our analysis though, we define per-step loss for language modeling, $l(p_{\theta}, w_{0}^{t-1}; o)$ as:
\begin{align}\label{eq:l_def}
  l(p_{\theta}, w_{0}^{t-1}; o) = \underset{w_{t} \sim o(\cdot|w_0^{t-1})}{\mathbb{E}} \log \frac{o(w_{t}|w_0^{t-1})}{p_{\theta}(w_{t}|w_0^{t-1})}, 
\end{align}
This definition ensures that the per-step loss for the oracle is zero; i.e., $l(o, w_{0}^{t-1}; o) = 0$. 

The per-step loss function defined by equation \ref{eq:l_def} ensures that the behavior cloning loss, $L^{BC}(p)$, under our definition is equivalent to teacher forcing loss, $L^{TF}(p)$, up to a constant term. This equivalence of $L^{BC}(p)$ and $L^{TF}(p)$ ensures that the model learned by minimizing either of the two losses will be identical.

% Substituting $l(p_{\theta}, w_{0}^{t-1}; o)$, from Equation \ref{eq:l_lm_def}, in Equation \ref{eq:l_pi_star_def}, we can define the behavior cloning loss for language modeling, $L^{BC}(p_{\theta})$ as:
% \begin{eqnarray}
%   \hspace*{-0.8em} L^{BC}(p_{\theta}) = \sum_{t=1}^{T}\underset{\substack{w_0^{t-1} \sim d_{o}^{t} \\ w_{t} \sim o(\cdot|w_0^{t-1})}}{\mathbb{E}} \log \frac{o(w_{t}|w_{0}^{t-1})}{p(w_{t}|w_{0}^{t-1})}. \label{eq:lm_lbc_def} \\
%   % \epsilon_t = \underset{\substack{w_0^{t-1} \sim d_{o}^{t} \\ w_{t} \sim o(\cdot|w_0^{t-1})}}{\mathbb{E}} \log \frac{o(w_{t}|w_{0}^{t-1})}{p_{\theta}(w_{t}|w_{0}^{t-1})}, \\
%   % = \sum_{t=1}^{T} \underset{\substack{w_0^{t-1} \sim d_{o}^{t}}}{\mathbb{E}} \text{KL}(o(\cdot |w_{0}^{t-1}) || p_{\theta}(\cdot|w_{0}^{t-1})). \nonumber \\ 
%   \approx \frac{1}{|\mathcal{D}|} \sum_{(w_{0}^{t-1}, w_i) \in \mathcal{D}} \log \frac{o(w_{t}|w_{0}^{t-1})}{p_{\theta}(w_{t}|w_{0}^{t-1})}, \label{eq:lm_lbc_emp_estimate}
% \end{eqnarray}
% where Equation \ref{eq:lm_lbc_emp_estimate} is the empirical estimate of $L^{BC}(p_{\theta})$ computed on the training data $\mathcal{D}$.

% We can relate maximum likelihood loss from Equation \ref{eq:tf_loss_def_1} to the behavior cloning loss (Equation \ref{eq:l_pi_star_def}) by defining the imitation loss $l(p_{\theta}, w_{0}^{t-1}; o)$ as:
% \begin{align}\label{eq:l_lm_def}
%   \hspace*{-0.8em} l(p_{\theta}, w_{0}^{t-1}; o) = \underset{w_{t} \sim o(\cdot|w_0^{t-1})}{\mathbb{E}} \log \frac{o(w_{t}|w_0^{t-1})}{p_{\theta}(w_{t}|w_0^{t-1})},
% \end{align}

\paragraph{Language Generation is Policy Rollouts:}
During policy rollouts, an agent in state $s_t$ executes the action $a_t$, sampled from policy $\pi$, and ends up in state $s_{t+1}$. The agent's next state is dependent upon its own actions. This state evolution can be formulated as sampling from state-visitation distribution induced by the policy $\pi$, i.e., $s_{t+1} \sim d_{\pi}^{t+1}$.

The performance of policy $\pi$ during rollouts can be measured using the loss (cost) of executing the policy $\pi$:
\begin{align} \label{eq:l_inference_def}
  \hspace*{-1em} L^{I}(\pi) =& \sum_{t=1}^{T} \mathbb{E}_{s_t \sim d_{\pi}^t} \left[l(s_t,\pi;o)\right] 
\end{align}

We can also formulate language generation in terms of policy rollouts from imitation learning. Mathematically, we can express generation as sampling contexts from the model's context distribution, i.e., $w_{0}^{j-1} \sim d_{p_{\theta}, \mathcal{F}}^{j}$, and generating the next token $w_{j}$ conditioned on $w_{0}^{j-1}$, using the decoding algorithm $\mathcal{F}$:
\begin{equation} \label{eq:lm_generation_def}
  \{w_{j} = \mathcal{F}(p_\theta, w_{0}^{j-1}) | w_{0}^{j-1} \sim d_{p_{\theta}, \mathcal{F}}^{j}\} 
\end{equation}
We can now define the inference-time loss, $L^{I}(p_{\theta})$, as the accumulated $T$-step loss of model $p_{\theta}$ imitating oracle $o$ on the context distribution induced by the model:
\begin{align}
  \hspace*{-0.6em} L^{I}(p_{\theta}) &= \sum_{t=1}^{T}\underset{\substack{w_0^{t-1} \sim d_{p_{\theta}, \mathcal{F}}^{t} \\ w_{t} \sim o(\cdot|w_0^{t-1})}}{\mathbb{E}} \log \frac{o(w_{t}|w_{0}^{t-1})}{p_{\theta}(w_{t}|w_{0}^{t-1})}, \label{eq:lm_li_kl_def} 
\end{align}
where $d_{p_{\theta}, \mathcal{F}}^{t}(w_{0}^{t}) \coloneqq p_{\theta}(w_{0}^{t-1})$, is the context distribution at step $t$, induced due to use of model $p_{\theta}$ and the decoding algorithm $\mathcal{F}$, from step $1$ to $t-1$. 

% \begin{align}
%   \hspace*{-0.8em} L^{I}(p_{\theta}) 
%   % &=  \sum_{t=1}^{T}\underset{w_0^{t-1} \sim d_{p_{\theta}, \mathcal{F}}^{t}}{\mathbb{E}} l(p_{\theta}, w_{0}^{t-1}; o). & \nonumber \\
%     &= \sum_{t=1}^{T}\underset{\substack{w_0^{t-1} \sim d_{p_{\theta}, \mathcal{F}}^{t} \\ w_{t} \sim o(\cdot|w_0^{t-1})}}{\mathbb{E}} \log \frac{o(w_{t}|w_{0}^{t-1})}{p_{\theta}(w_{t}|w_{0}^{t-1})}, & \label{eq:lm_li_kl_def} %\\
%     % &= \sum_{t=1}^{T} \underset{\substack{w_0^{t-1} \sim d_{p_{\theta}, \mathcal{F}}^{t}}}{\mathbb{E}} \text{KL}(o(\cdot |w_{0}^{t-1}) || p_{\theta}(\cdot|w_{0}^{t-1})), &  \nonumber 
% \end{align}

\section{Exposure Bias and Error Accumulation}~\label{sec:qeb_err_acc}
\citet{ranzato_sequence_2015} defined exposure bias as a behavioral mismatch between maximum likelihood-based training and the generation procedure at inference time. During maximum likelihood-based training, the next token distribution is conditioned on ground truth data whereas, during generation, it has to rely on the model's own previously generated tokens. They also postulated that this training and generation context distribution mismatch might result in an accumulation of errors during generation. 

Intuitively, when the model produces a token $w_i$ that makes the resulting context $w_0^{i}$ unfamiliar, it might not be able to continue the generation adequately and is likely to produce another token which will further make the context flawed. This phenomenon reinforces itself as the context drifts further from what the oracle would produce, leading to an accumulation of errors.

In the imitation learning literature, the accumulation of errors while rolling out a policy trained using behavior cloning is analyzed in the terms of inference-time regret of the behavior cloning policy, $\pi_{BC}$, with respect to the oracle policy, $o$,~\cite{ross_efficient_2010,ross_reduction_2011} i.e.,
\begin{equation}\label{eq:il_regret}
  \mathcal{R}(\pi_{BC}) = L^I(\pi_{BC}) - L^I(o) 
\end{equation}
Let $\epsilon_{t}$ be the expected error of executing policy $\pi$ at step $t$ on the state-visitation distribution induced by the oracle $o$, i.e., 
\begin{equation}
  \epsilon_{t} = \mathbb{E}_{s \sim d^{t}_{o}} [l(s, {\pi}; o)]
\end{equation}
Let $\epsilon$ be the average expected error of executing policy $\pi$ over $T$ step, i.e.,  $\epsilon = 1/T \sum_{t=1}^{T} \epsilon_{t}$. Assuming $l(s,\pi, o)$ is an upper bound on $[0, 1]$ loss, we can bound the regret for a policy ${\pi}_{BC}$ as,
\begin{equation} \label{eq:il_quad_growth}
  T\epsilon \le \mathcal{R}(\pi_{BC}) \le T^{2}\epsilon.
\end{equation}
The lower bound in Equation~\ref{eq:il_quad_growth} assumes no accumulation of error, hence an expected error of $\epsilon$ at each step, whereas the upper bound assumes the worst-case scenario, resulting in linear growth in error at each step and overall quadratic accumulative growth w.r.t. maximum sequence length $T$.

% The possible super-linear growth of the error is caused by the fact that a prediction error might lead the policy into a state that was either infrequently or never encountered during training. This might lead to more errors with the worst-case scenario of committing an error of cost $u$ at each time step.

Relying on the imitation learning perspective of language generation presented in the previous section, we can now borrow this regret-based analysis from imitation learning literature to similarly bound the regret of a language generation model as
\begin{equation} \label{eq:lm_quad_growth}
  T\epsilon \le \mathcal{R}(p_{\theta}, \mathcal{F}) \le T^{2}\epsilon.
\end{equation}
where $p_{\theta}$ is the model being used for generation,  $\mathcal{F}$ is the decoding method being used for generation, $\epsilon =  1/T \sum_{t=1}^{T} \epsilon_{t}$ and $\epsilon_{t}$ is defined as 
\begin{equation}
  \epsilon_{t} = \underset{\substack{w_0^{t-1} \sim d_{o}^{t} \\ w_{t} \sim o(\cdot|w_0^{t-1})}}{\mathbb{E}} \log \frac{o(w_{t}|w_{0}^{t-1})}{p_{\theta}(w_{t}|w_{0}^{t-1})}
\end{equation}
We will now use these bounds on the regret to analyze and quantify the error accumulation due to exposure bias in language generation. 

\section{Quantifying Error Accumulation due to Exposure Bias}\label{sec:quantify_exp_bias}

In our analysis, we use two metrics, $\text{AccErr}_{\le}(l)$ and $\text{\%ExAccErr}_{\le}(l)$ to measure the impact of error accumulation due to exposure bias.

We define accumulated errors up to length $l$, $\text{AccErr}_{\le}(l)$, as 
\begin{equation}
  \text{AccErr}_{\le}(l) = \mathcal{R}_{\le l}(p_{\theta}, \mathcal{F})/\epsilon_{\le l}
\end{equation}
Here, $\mathcal{R}_{\le l}(p_{\theta}, \mathcal{F})$ be the regret due to the use of language model $p_{\theta}$ and decoding method, $\mathcal{F}$, up to sequence length $l$, and $\epsilon_{\le l} = 1/l \sum_{t=1}^{l}\epsilon_{t}$ is the expected per-step error up to length $l$.

% Let $\epsilon_{t}$, be the expected per-step loss over the context distribution induced by the oracle $o$, i.e., 
% \begin{multline*} \label{eq:lg_epsilon_def}
%   \mathcal{R}(p_{\theta},  \mathcal{F})  = \sum_{t=1}^{T}\underset{\substack{w_0^{t-1} \sim d_{p_{\theta}, \mathcal{F}}^{t} \\ w_{t} \sim o(\cdot|w_0^{t-1})}}{\mathbb{E}} \log \frac{o(w_{t}|w_{0}^{t-1})}{p_{\theta}(w_{t}|w_{0}^{t-1})}, \\
%   = \sum_{t=1}^{T} \underset{\substack{w_0^{t-1} \sim d_{p_{\theta}, \mathcal{F}}^{t}}}{\mathbb{E}} \text{KL}(o(\cdot |w_{0}^{t-1}) || p_{\theta}(\cdot|w_{0}^{t-1})), \\
% \end{multline*}
% Let  We know from Equation \ref{eq:il_quad_growth}, that the $\mathcal{R}_{\le l}(p_{\theta}, \mathcal{F})$ is bounded by $l\epsilon_{\le l}$ on the left side, i.e., 
% \begin{equation}  \label{eq:lm_super_linear_growth}
%   \mathcal{R}_{\le l}(p_{\theta}, \mathcal{F}) \ge l\epsilon_{\le l}
% \end{equation}
% This inequality allows us to analyze the error accumulation due to exposure bias, i.e., if $\mathcal{R}_{\le l}(p_{\theta}, \mathcal{F})$ grows faster than $l\epsilon_{\le l}$, it can be inferred that the language generation under model $p_{\theta}$, and decoding algorithm, $\mathcal{F}$ suffers from an accumulation of errors due to exposure bias.

This metric captures the growth of error w.r.t. sequence length $l$. \textul{If  exposure bias does indeed leads to error accumulation, $\text{AccErr}_{\le}(l)$ should grow super-linearly w.r.t. $l$.}

We define our second metric, $\text{\%ExAccErr}_{\le}(l)$, as the percentage of excess errors committed by the model that can be attributed to exposure bias, i.e.,
\begin{equation*}
  \text{\%ExAccErr}_{\le}(l) = \frac{\mathcal{R}_{\le l}(p_{\theta}, \mathcal{F}) - l\epsilon_{\le l}}{l\epsilon_{\le l}} * 100
\end{equation*}
Here, $l\epsilon_{\le l}$ is the lower bound on the regret and is the minimum number of errors ($\epsilon$ per step) a model would make if there was no accumulation of errors.

$\text{\%ExAccErr}_{\le}(l)$ allows us to compare models, training algorithms, and decoding strategies on the extra error that might be caused/mitigated by their use. \textul{A model, training algorithm, or decoding strategy that perfectly mitigates the exposure bias will result in zero excess accumulated errors.}

In the rest of the paper, we use these definitions to show: 1) error accumulation in language generation is real, 2) perplexity fails to capture this error accumulation, 3) lower exposure bias correlates with a higher quality generation that is more coherent, uses more diverse vocabulary, and is less repetitive. 

% In the rest of the paper, we empirically validate the theoretical results from the previous section and show that errors accumulate due to exposure bias and these errors grow nearly quadratically. We will also use our analysis to show that perplexity can only measure the per-step generalization error gap and fails to capture the error accumulation due to exposure bias. Finally, we evaluate the impact of decoding algorithms on exposure bias and show that decoding algorithms with lower exposure bias also result in a higher quality generation. 

% \begin{figure}[tbp!]
%   \centering
%       \includegraphics[width=0.5\textwidth]{img/acc_err.png}
%     \caption{\textcolor{blue}{$AccErr_{\le}(l)$} vs \textcolor{orange}{$l$}: This graph shows the quadratic growth of accumulated errors w.r.t to sequence length ($l$) as predicted by the theory.}
%     \label{fig:acc_err_till_len_l}
% \end{figure}
% \begin{figure}[tbp!]
%   \centering
%       \includegraphics[width=0.5\textwidth]{img/excess_err.png}
%     \caption{\textcolor{blue}{$\text{\%ExError}_{\le}(l)$} vs $l$: \% excess errors due to error accumulation caused by exposure bias. This indicates that \textbf{extra errors} due to exposure bias grows near-linearly with the sequence length, and decoding using greedy search results in over 70\% more errors.}
%     \label{fig:pct_excess_err_till_l}
% \end{figure}

\begin{figure*}[tbp!]
  \begin{subfigure}[b]{0.49\textwidth}
    \includegraphics[width=1\textwidth]{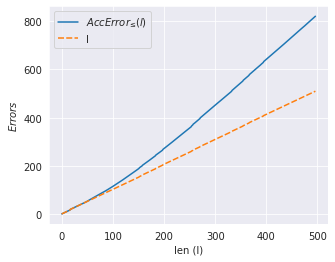}
    \caption{\textcolor{blue}{$AccErr_{\le}(l)$} vs \textcolor{orange}{$l$}}
    \label{fig:acc_err_till_len_l}
  \end{subfigure}
  \begin{subfigure}[b]{0.48\textwidth}
      \includegraphics[width=1\textwidth]{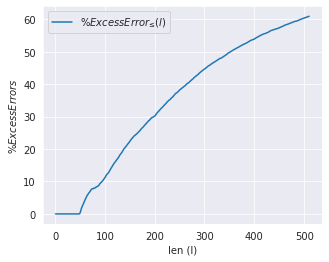}
      \caption{\textcolor{blue}{$\text{\%ExError}_{\le}(l)$} vs $l$.}
      \label{fig:pct_excess_err_till_l}
    \end{subfigure}
  \label{fig:}
  \caption{Figure \ref{fig:acc_err_till_len_l} plots accumulated error till length $l$ (\textcolor{blue}{$AccErr_{\le}(l))$}) w.r.t. \textcolor{orange}{$l$}. This graph shows the quadratic growth of accumulated errors w.r.t to sequence length ($l$) as predicted by the theory. Figure \ref{fig:pct_excess_err_till_l} plots \% excess errors due to error accumulation (\textcolor{blue}{$\text{\%ExError}_{\le}(l)$}) caused by exposure bias. This indicates that \textbf{extra errors} due to exposure bias grows near-linearly with the sequence length, and decoding using greedy search results in over 70\% more errors.} 
\end{figure*}

\section{Study Setup: Open-ended Generation}
\paragraph{Text Completion Setup:} Text completion is the standard experimental setup to measure the quality of generation in open-ended language generation \cite{holtzman_curious_2019,welleck_neural_2019}. It is also a generalization of numerous practical language generation applications such as story generation~\cite{fan_hierarchical_2018}, contextual text completion~\cite{radford_language_nodate}, dialog modeling~\cite{zhang_composable_2018}. 

 Text completion models take a text passage or prefix $w_{0}^{j} \sim o$ as an input and generate a coherent continuation of the prefix, $w_{j+1}^{n}$ using the language model $p_{\theta}$ and the decoding algorithm $\mathcal{F}$, i.e., $w_{j+1}^{n} = \mathcal{F}(p_{\theta}, w_{0}^{j})$. In this paper, we use this text-completion setup to analyze the error accumulation due to exposure bias and its correlation with language generation quality.

\paragraph{Language Model and Dataset:}
We conduct our analysis using the GPT2 language model~\cite{radford_language_nodate}. We use the GPT2-117M model as our evaluation language model and use the train split of Wikitext-103~\cite{merity_pointer_2016} for prompts. We rely on GPT-2 model fine-tuned on Wikitext-103 as our approximate oracle. We tokenize the Wikitext-103 dataset using GPT-2's tokenization scheme. We chunk Wikitext-103's train split into sequences of length $512$.  Of these, we use the first $50$ tokens as prompts for our generation experiments and generate the completions to a maximum length of $512$ or up to the end of the sequence token. We use a total of $20k$ prompts for our evaluation. 

\section{Results}
\subsection{Error Accumulation in Language Generation is Real!}
% \paragraph{Accumulation w.r.t. Sequence Length}
Figure \ref{fig:acc_err_till_len_l} plots $\text{AccErr}_{\le}(l)$ w.r.t. sequence length, $l$. The support (dotted, orange line) $y=x$, captures the linear growth.  It shows $\text{AccErr}_{\le}(l)$ grows near-quadratically w.r.t. sequence length, empirically validating the theory that exposure bias would lead to the accumulation of errors. Figure \ref{fig:pct_excess_err_till_l}, further strengthens this claim by demonstrating near-linear growth in excess errors w.r.t. to the sequence length. 

We hypothesize that these excess errors would manifest in the form of language degeneration, especially in the latter part of the sequence, and would cause issues such as hallucinations, limited vocabulary, and word- and phrase-level repetitions. % Some qualitative evidence of this phenomenon is provided in Table~\ref{table:decoding_vs_exposure_bias_qualitative}, especially for samples generated using greedy and beam search decoding schemes.

% \paragraph{Accumulation w.r.t. Context Length}

% \begin{figure}[tbp!]
%   \centering
%       \includegraphics[width=0.5\textwidth,height=0.25\textheight]{img/context_len_vs_exposure_bias.png}
%     \caption{Error Accumulation w.r.t. Context Length}
%     \label{fig:}
% \end{figure}

\begin{figure}[t!]
  \centering
    \includegraphics[width=0.5\textwidth]{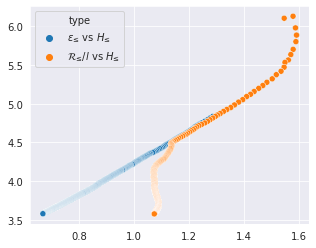}
    % \caption{\textcolor{blue}{$\epsilon_{\le l}$ vs $L^{h}_{\le l}$. ($\rho=0.9997$)} and \textcolor{orange}{$R^{\le l}(p_{\theta})/l$ vs  $L^{h}_{\le l}$. ($\rho=0.4003$)}}
  \caption{Analyzing (log) perplexity ($H_{\le l}$) w.r.t to average per-step error ($\epsilon_{\le l}$), and length-normalized exposure bias regret ($\mathcal{R}_{\le l}(p_{\theta}, \mathcal{F})/l$). We observe that perplexity strongly correlates with average per-step error ($\rho=0.9997$), but it has a weaker correlation with length-normalized regret ($\rho=0.4003$).}
  \label{fig:epsilon_vs_ppl}
\end{figure}

\begin{table*}[t!]
	\centering \ra{1.1}
	\begin{tabular}{@{}lcccccc@{}}
    \toprule
		\multirow{ 2}{*}{Search} & & \multicolumn{4}{c}{Generation Quality} \\
    \cmidrule{3-7}
                  & \%ExErrAcc ($\downarrow$) &  seq-rep-4 ($\downarrow$) & rep ($\downarrow$) & wrep ($\downarrow$) & uniq ($\uparrow$) \\ 
    \midrule
		Greedy & 60.96\%  & 0.8990 & 0.4423 & 0.4136 & 7833 \\ 
		Beam (k=5) & 69.72\%  & 0.8094 & 0.4064 & 0.3787 & 10966 \\
    \hdashline
    \multicolumn{7}{@{}l}{Sampling} \\
    \hspace{0.5em} w/ Temp (temp=1)     & 39.37\%  & 0.1883 & 0.2547 & 0.2301 & 23729 \\
		 \hspace{0.5em} w/ Temp (temp=1.2)  & \textbf{24.75\%}  & \textbf{0.1556} & \textbf{0.2271} & \textbf{0.2033} & 25225 \\
     \hspace{0.5em} w/ top-k (k=100)    & 35.37\%  & 0.1690 & 0.2409 & 0.2166 & \textbf{26251} \\
     \hspace{0.5em} w/ top-p (p=0.94)   & 48.71\%  & 0.2218 & 0.2743 & 0.2490 & 22582 \\
    \hline 
    Human                               & -        & 0.0274 & 0.4338 & - & 28739 \\ 
    \bottomrule
	\end{tabular}
	\caption{Impact of error accumulation on generation quality. We observe that stochastic decoding methods not only lead to diverse language generation but also have lower exposure bias than the deterministic methods.}
  \label{table:decoding_vs_exposure_bias}
\end{table*}

\subsection{Perplexity is Not Enough}\label{sec:perplexity_is_not_enough}
Perplexity is a standard measure used to evaluate the quality of a language model. It is often used as a proxy measure for the text generation quality of the language model. In this section, we argue perplexity paints an incomplete picture regarding a model's ability to generate high-quality, coherent text. It only captures the average per-step error generalization gap (or lack of it) but fails to account for the error accumulation due to exposure bias. These accumulated errors, as seen in the previous section, can grow near-quadratically and can prove to be a major concern for any generation model that generates sequences longer than a few words.

Perplexity can be seen as scaled exponentiated average per-step error, $\epsilon$, computed over a held-out test set,  $\mathcal{D}_{h}$:
\begin{align}
  \epsilon  & = 1/T\sum_{t=1}^{T}\underset{\substack{w_0^{t-1} \sim d_{o}^{t} \\ w_{t} \sim o(\cdot|w_0^{t-1})}}{\mathbb{E}} \log \frac{o(w_{t}|w_{0}^{t-1})}{p(w_{t}|w_{0}^{t-1})}.  \\
    & \approx \frac{-1}{|\mathcal{D}_{h}|} \sum_{(w_{0}^{i-1}, w_i) \in \mathcal{D}_{h}} \log p_{\theta}(w_{i}|w_{0}^{i-1}) + c, \\
    & = H(p_{\theta}; \mathcal{D}_{h}) + c. \label{eq:perplexity_per_step_error}
` '\end{align}
where $H(p_{\theta}; \mathcal{D}_{h})$ is the entropy rate (log perplexity) of the model $p_{\theta}$ on the held-out test set $\mathcal{D}_{h}$.

As entropy rate is a linear function of average per-step error, we hypothesize that it will only be able to measure the per-step generalization gap of the model and will fail to capture the error accumulation caused by reducing a sequential decision-making problem to a supervised learning problem. 

In Figure \ref{fig:epsilon_vs_ppl}, we plot the entropy rate, $H(p_{\theta}; \mathcal{D}_{h})_{\le l}$, w.r.t. average per-step error, $\epsilon_{\le l}$, and length-normalized regret up to length $l$, $\mathcal{R}_{\le l}(p_{\theta}, \mathcal{F})/l$. We observe a strong correlation between the entropy rate and average per-step error ($\rho=0.9997$) validating our theoretical observation that perplexity can capture the per-step generalization gap of language model $p_{\theta}$. On the other hand, the length-normalized regret exhibits a poor correlation with the entropy rate ($\rho=0.4003$) indicating perplexity's failure to capture the error accumulation due to exposure bias.

A case in point of perplexity's inability to capture error accumulation is the degenerate behavior of GPT-2~\cite{radford_language_nodate} while generating moderately long sequences under greedy or beam search. This happens despite GPT2 having a low zero-shot perplexity on the held-out set of the Wikitext-103 dataset (perplexity: $37.50$). We hypothesize that one of the reasons for the degenerate behaviors of large pre-trained language models such as repetition, low vocabulary usage, and a lack of coherent generation is the result of this accumulation of errors. An example of this behavior is presented in Table~\ref{table:decoding_vs_exposure_bias_qualitative} where we observe GPT2 generating repetitive and incoherent text completion for a WikiText-103 prompt under deterministic decoding schemes such as greedy and beam decoding.

\subsection{Error Accumulation Impacts Generation Quality}

\begin{table}[t!] \ra{1.5}
  \small
	\begin{tabular}{@{}p{0.48\textwidth}@{}}
		\textbf{Context}: ... He made one attempt to break through the blockade on 23 December, but it proved impossible to do as the Union controlled every channel from Savannah, aided ...\\ \hdashline
		\textbf{Greedy}: ... Savannah, aided by the British, to the North.  The blockade was broken on 23 December by the British and the Americans. \textcolor{red}{The blockade was broken on 23 December by the British and the Americans.} \textcolor{red}{The blockade was broken on 23 December by the British and the Americans.} \textcolor{red}{The blockade was broken on 23 December by the British and the} ...\\ \hline
		% \textbf{Greedy w/ w/ repetition penalty} (pen=1.2): \\ \hline
		\textbf{BeamSearch} (k=5): ... Savannah, aided by the North Carolina Railroad. In the spring of 1788, in the midst of the blockade, he made a move to join the Confederacy, \textcolor{red}{but it was too late.}  On 23 December 1788, he marched to the battle of Custer's Creek in the South Carolina River.  In the following weeks, he returned to Savannah \textcolor{red}{on 23 December, but it was too late.}  On 24 December 1788, he made another attempt to break through the blockade \textcolor{red}{on 23 December, but it was too late.} ...\\ \hline
		\textbf{Sampling w/ Temp.} (temp=1.2): ... Savannah, aided by US Navy ships, and took him by surprise. One of the most famous incidents involved General Grant's arrest at the Battle of Camp Amish on 16 August. He was being held by the Union Army in Atlanta and accused of taking down General Joseph E. Douglas, a Marine commander who was in command of Fort Hood, an Army outpost near Fort Laramie, Mississippi. ...\\ \hline
		\textbf{Top-k Sampling} (k=100): ... Savannah, aided by a little naval warplane bombardment which had crippled many parts of the ship. As an early American merchant vessel he was carried forward as a heavy merchant, while at some time it was necessary to take up residence in Savannah to be used as a regular transport vessel. In the meantime, on the 20th of January 1816, Jefferson died. He was buried in an unmarked grave in the chapel inside Savannah in the family home at Glenbrook, where the rest of the family resides. ...\\ \hline
		\textbf{Top-p Sampling} (p=0.94): ... Savannah, aided in the capture of his land and his supplies. He made two attempts to stop a blockade on 28 December, but failed to make it. There were, however, six people in Savannah who were convinced that Captain Walker was the cause of the blockade. He then made a trip to North Carolina where he gave up hope. ...\\ \hdashline
		\textbf{Gold}: ... He made one attempt to break through the blockade on 23 December, but it proved impossible to do as the Union controlled every channel from Savannah, aided by their occupation of Tybee Island at the mouth of the Savannah River. Bulloch reported to Mallory in late January 1862 that breaking out was hopeless so Mallory ordered him to turn the ship over to another officer and to return to Europe some other way. ... \\ 
	\end{tabular}
	\caption{Examples of completions using various decoding methods. We observe that the deterministic decoding schemes produce less diverse, incoherent, and more repetitive (highlighted in \textcolor{red}{red}) text.}
  \label{table:decoding_vs_exposure_bias_qualitative}
\end{table}

Finally, we examine the hypothesis that poor text generation capabilities of pre-trained large language models under greedy decoding might be due to the error accumulation caused by a procedural mismatch between generation and maximum likelihood training~\citep{vijayakumar_diverse_2016,welleck_neural_2019,holtzman_curious_2019}.

The regret-based definition of error accumulation allows us to analyze exposure bias along two axes of variation: the trained language model, $p_{\theta}$, and the decoding algorithm, $\mathcal{F}$. In this set of experiments, we explore the impact of various decoding schemes on error accumulation due to exposure bias and the quality of the completed text. 

For a quantitative analysis of the impact of various decoding algorithms on the quality of language generation, we measure the completion quality by using the same metrics as \citet{welleck_neural_2019}. These metrics are: 1.) \textbf{rep}/128 measures if the prediction token at step $t$ occurs in previous $128$ steps, 2.) \textbf{wrep}/128 counts the prediction's repetition at step $t$ only if the predicted token is not the ground-truth token at that position, 3.) \textbf{seq-rep-4} measure the repetition at the 4-gram level, and 4.) \textbf{uniq} measure the vocabulary diversity by accounting for the number of unique tokens generated by the model.

Table \ref{table:decoding_vs_exposure_bias} shows that various sampling-based decoding algorithms result in diverse and more coherent language generation and a lower percentage of excess errors. Sampling with temperature (with temp=$1.2$) leads to the least amount of repetition (both at the token and the n-gram level), second highest vocabulary diversity, and the least amount of excess errors due to exposure bias. This also bears out from our qualitative analysis in Table \ref{table:decoding_vs_exposure_bias_qualitative} as sampling with temperature produces the most coherent text. Greedy and beam search decoding schemes, in contrast, fare poorly in both reducing exposure bias and language generation quality metrics, producing repetitive and incoherent text. 
These quantitative and qualitative experiments offer us evidence that reducing exposure bias does lead to more coherent text generation.

We hypothesize that the reasonable amount of randomness introduced by stochastic sampling helps the model avoid sampling the most likely token at each time step, thus avoiding possible divergent contexts that might have resulted in a degenerate completion in the future. We conjecture that this timely intervention prevents the generation context distribution from diverging too far away from the training context distribution, helping it avoid the compounding of errors. This is also borne out by qualitative analysis as a reasonable amount of stochasticity does result in texts which look more coherent and oracle-like. A broader analysis of this behavior though is beyond the scope of this work and is left for future work.

% Benchmarking various decoding strategies allows us to verify if an accumulation of errors does indeed lead to degeneration as the choice of decoding algorithm does not impact the average per-step error, $\epsilon$, or held-out test set perplexity ($H(p_{\theta}; \mathcal{D}_{h})_{\le l}$). This rules out the role of modeling and model training in language degeneration across different decoding algorithms. Hence, it is reasonable to causally link a decoding algorithm's resultant improvement in language generation to its ability to reduce error accumulation.

Considering that choice of decoding algorithm does not impact average per-step error, $\epsilon$, this rules out the role of modeling and model training in language degeneration. Hence, it is reasonable to assume that both qualitative and quantitative improvement in language quality observed in this experiment is strongly linked to the reduction in error accumulation due to exposure bias.

\section{Related Work}\label{sec:related_work}

\textbf{Non-MLE Training Methods}: 
Several approaches have been proposed to mitigate the exposure bias issue including RL-based optimization objectives \citep{ranzato_sequence_2015,shen_minimum_2016,bahdanauActorCritic2017,chen_reinforcement_2020}, learning to search \citep{leblond_searnn:_2017}, energy-based models \citep{deng_residual_2020}, imitation learning \citep{du_empirical_2019}, generative adversarial networks \citep{yu_seqgan:_2016} and knowledge distillation \citep{liu_teacher-student_2020}. Although these methods motivate their approaches as intending to reduce exposure bias, they neither formally analyze exposure bias nor provide any empirical evidence that these methods mitigate the effect of exposure bias. In this paper, we analyze the exposure bias from a principled imitation learning perspective in terms of the accumulation of errors. This definition can be adapted to evaluate various novel training and modeling approaches on their ability to reduce exposure bias.

\textbf{Smarter Decoding Methods}: Large language models have unusually low test perplexities but they falter at coherent and diverse language generation specifically in open-ended language generation tasks especially while using deterministic decoding schemes. Several authors~\citep{vijayakumar_diverse_2016,welleck_neural_2019,holtzman_curious_2019} have hypothesized that training and inference mismatch due to MLE-based training is responsible for the degenerate behavior. They have proposed smarter decoding schemes to mitigate the side effects of exposure bias resulting in better generation quality. Despite this being an active area of research, this often-repeated hypothesis for degenerate generation behavior has not received serious treatment until now. In this paper, we take a step towards explaining this discrepancy and show that error accumulation due to exposure bias might be the reason for this degenerate behavior and explain why perplexity has a handicap in capturing this compounding of errors.

\textbf{Analyzing Exposure Bias}: 
\citet{schmidt_generalization_2019} and \citet{wang_exposure_2020} link exposure bias to a generalization gap due to distribution and domain shift respectively. Performance degradation under domain and distribution shift is a major issue with language generation, and direct evidence supporting this hypothesis will provide insights into building more robust language generation models. Unfortunately, neither of the papers formally analyzes the exposure bias issue or empirically links the generalization gap to exposure bias directly.

% Specifically, \citet{wang_exposure_2020} shows that minimum risk training (MRT)---a non-MLE training objective, performs better under domain shift and produces fewer hallucinations. The authors attribute MRT's robustness to domain shift to its ability to address the exposure bias issue.
% \citet{schmidt_generalization_2019} analyzes the connection between exposure bias and a model's ability to generalize beyond its training data and highlights perplexity's (test likelihood's) inability to capture the generalization gap due to exposure bias. \citet{wang_exposure_2020} furthers the generalization gap perspective of exposure bias by showing that minimum risk training (MRT)---a non-MLE training objective, performs better under domain shift and produces fewer hallucinations. The authors attribute MRT's robustness to domain shift to its ability to redress the exposure bias issue. Notably, neither of these papers formalize the notion of exposure bias or empirically link the generalization gap to exposure bias. 

Three recent papers, \citet{xu_rethinking_2020,zhang_bridging_2019,he_quantifying_2020}, have tried to empirically evaluate the impact of exposure bias on language generation. The first two papers validate the existence of exposure bias whereas \citet{he_quantifying_2020} show language models have self-recovering ability negating the impact of exposure bias. All three analyses are based on the empirical definition of exposure bias which, in turn, is based on the informal formulation by \citet{ranzato_sequence_2015}. 

In this paper, we provide a principled and theoretically grounded approach to analyze exposure bias in language generation and show that it is indeed a problem and that it might explain the degeneration issue with large language models on open-ended tasks under deterministic decoding.

\section{Discussion}
% Inspired by https://twitter.com/iankleckner/status/1458433895876153355?s=20.

  In this paper, we analyze language generation from an imitation learning perspective. We use this analysis to arrive at a theoretical bound on error accumulation due to exposure bias. This bound predicts a super-linear growth in error accumulation during generation due to exposure bias. In our experiments, we validate this bound and show that accumulation due to exposure bias indeed results in super-linear growth in errors. 

  We then show, both analytically and empirically, why perplexity is not enough to capture this accumulation of errors and hypothesize that this accumulation of errors is responsible for the degenerate language generation. Finally, we provide some evidence for this hypothesis by evaluating the impact of various decoding schemes on error accumulation and generation quality. We show that techniques that improve the generation quality do result in a lower error accumulation and this indicates that excess error accumulation due to exposure bias might be a factor affecting language generation quality.

  Our analysis provides a principled and theoretically grounded way to understand exposure bias. We believe this analysis can pave way for developing smarter training and decoding algorithms to address this error accumulation resulting in more robust language generation models. 
  
  % A possible limitation of this work is its reliance on fine-tuned GPT-2 model as an approximate oracle. We acknowledge these concerns, and justify this choice by stating that the oracle in our analysis was fine-tuned on the same data used for prompts. A possible extension of this work can be to analyze this choice and benchmark other possible oracle choices and verify the conclusions of our analysis under these choices. Another possible extension of this work is to analyze various training algorithms such as MRT~\cite{ochMinimumErrorRate2003,shen_minimum_2016}, SEARNN~\cite{leblond_searnn:_2017}, REINFORCE~\cite{ranzato_sequence_2015}, and Scheduled Sampling~\cite{bengio_scheduled_2015} on their ability to alleviate exposure bias issue.

\section*{Acknowledgments}
We would like to thank the reviewers for their valuable feedback. This work is supported by funding from Samsung Electronics. The last author
is supported by the Canada CIFAR AI Chair program. This research was enabled in part by support provided by Calcul Québec \footnote{\url{https://www.calculquebec.ca}}, and Compute Canada \footnote{\url{https://www.computecanada.ca}}. We would also like to thank Khimya Khetarpal, Sachin Grover, Ankit Anand, and Jayakumar Subramanian for feedback on current and previous drafts of this paper, and colleagues at Borealis AI for their valuable inputs and discussions during the first author's internship at Borealis AI.
\bibliography{emnlp2020,references_new}
\bibliographystyle{acl_natbib}

\end{document}